\documentclass[10pt,twocolumn,letterpaper]{article}

\usepackage{cvpr}              

\usepackage{graphicx}
\usepackage{amsmath}
\usepackage{amssymb}
\usepackage{booktabs}
\usepackage{comment}
\usepackage{multirow}
\usepackage{mathtools}
\usepackage{epsfig}
\usepackage[accsupp]{axessibility}

\usepackage[dvipsnames]{xcolor}
\usepackage[breaklinks,colorlinks,citecolor = ForestGreen]{hyperref}

\usepackage[capitalize]{cleveref}
\crefname{section}{Sec.}{Secs.}
\Crefname{section}{Section}{Sections}
\Crefname{table}{Table}{Tables}
\crefname{table}{Tab.}{Tabs.}

\def\cvprPaperID{3038} 
\def\confName{CVPR}
\def\confYear{2022}

\begin{document}

\title{PhysFormer: Facial Video-based Physiological Measurement with \\ Temporal Difference Transformer}

\author{Zitong Yu\textsuperscript{1}, Yuming Shen\textsuperscript{2}, Jingang Shi\textsuperscript{3}, Hengshuang Zhao\textsuperscript{4,2}, Philip Torr\textsuperscript{2}, Guoying Zhao\textsuperscript{1\thanks{Corresponding author}}\\
\textsuperscript{1}CMVS, University of Oulu  \qquad  \textsuperscript{2}TVG, University of Oxford  \\
\textsuperscript{3}Xi'an Jiaotong University  \qquad 
\textsuperscript{4}The University of Hong Kong 
}

\maketitle

\begin{abstract}
\vspace{-0.6em}
Remote photoplethysmography (rPPG), which aims at measuring heart activities and physiological signals from facial video without any contact, has great potential in many applications. Recent deep learning approaches focus on mining subtle rPPG clues using convolutional neural networks with limited spatio-temporal receptive fields, which neglect the long-range spatio-temporal perception and interaction for rPPG modeling. In this paper, we propose the PhysFormer, an end-to-end video transformer based architecture, to adaptively aggregate both local and global spatio-temporal features for rPPG representation enhancement. As key modules in PhysFormer, the temporal difference transformers first enhance the quasi-periodic rPPG features with temporal difference guided global attention, and then refine the local spatio-temporal representation against interference. Furthermore, we also propose the label distribution learning and a curriculum learning inspired dynamic constraint in frequency domain, which provide elaborate supervisions for PhysFormer and alleviate overfitting. Comprehensive experiments are performed on four benchmark datasets to show our superior performance on both intra- and cross-dataset testings. One highlight is that, unlike most transformer networks needed pretraining from large-scale datasets, the proposed PhysFormer can be easily trained from scratch on rPPG datasets, which makes it promising as a novel transformer baseline for the rPPG community. The codes are available at \href{https://github.com/ZitongYu/PhysFormer}{https://github.com/ZitongYu/PhysFormer}.

\end{abstract}

\vspace{-1.0em}
\section{Introduction}

\thispagestyle{empty}

\begin{figure}
\vspace{-0.6em}
\centering
\includegraphics[scale=0.45]{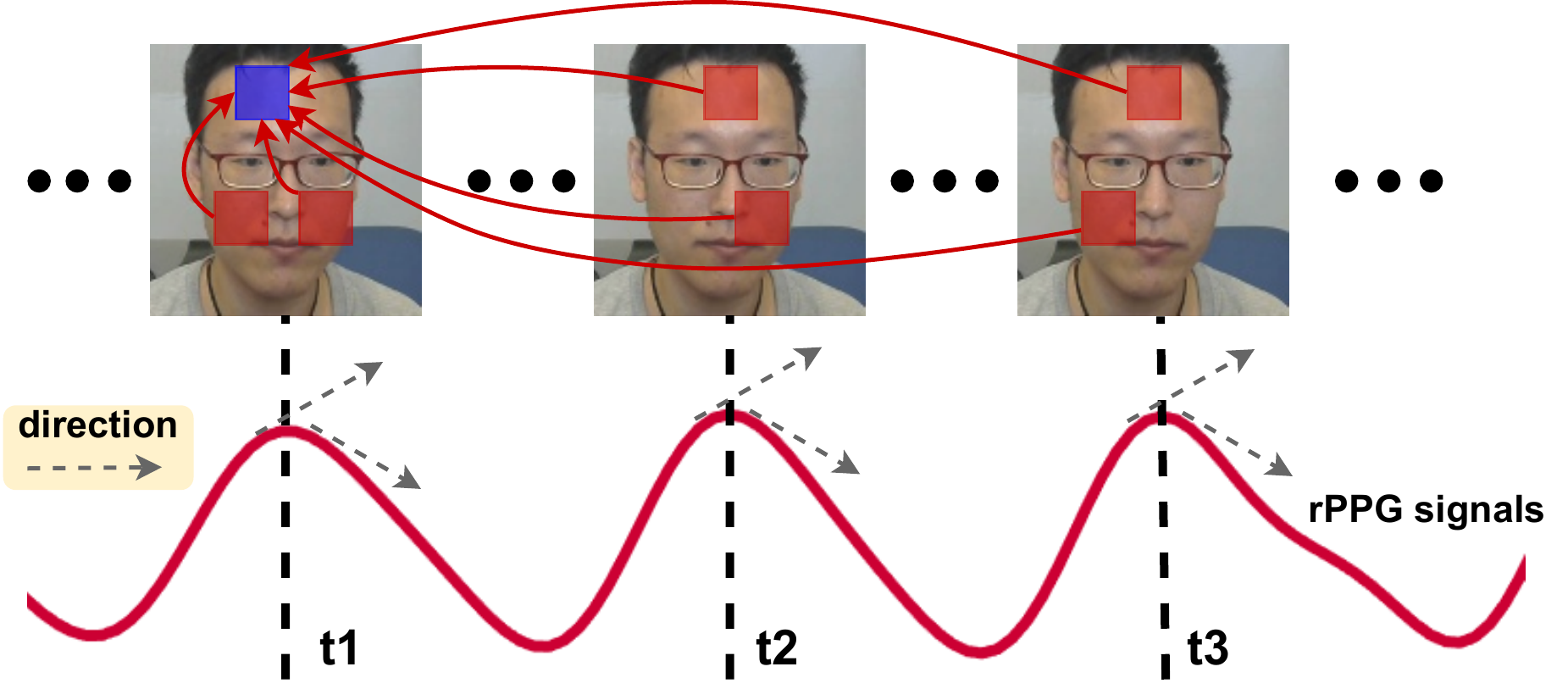}
\vspace{-1.7em}
  \caption{\small{
  The trajectories of rPPG signals around t1, t2, and t3 share similar properties (e.g., trends with rising edge first then falling edge later, and relatively high magnitudes) induced by skin color changes. It inspires the long-range spatio-temporal attention (e.g., \textcolor{blue}{blue} tube around t1 interacted with \textcolor{red}{red} tubes from intra- and inter-frames) according to their local temporal difference features for quasi-periodic rPPG enhancement. Here `tube' indicates the same regions across short-time consecutive frames.}
  }
 
\label{fig:Figure1}
\vspace{-0.7em}
\end{figure}

Physiological signals such as heart rate (HR), respiration frequency (RF), and heart rate variability (HRV) are important vital signs to be measured in many circumstances, especially for healthcare or medical purposes. Traditionally, the Electrocardiography (ECG) and Photoplethysmograph (PPG) are the two most common ways for measuring heart activities and corresponding physiological signals. However, both ECG and PPG sensors need to be attached to body parts, which may cause discomfort and are inconvenient for long-term monitoring. To counter for this issue, remote photoplethysmography (rPPG)~\cite{yu2021facial,chen2018video,liu2021camera} methods are developing fast in recent years, which aim to measure heart activity remotely without any contact.

In earlier studies of facial rPPG measurement, most methods analyze subtle color changes on facial regions of interest (ROI) with classical signal processing approaches~\cite{verkruysse2008remote,poh2010non,poh2010advancements,li2014remote,tulyakov2016self}. Besides, there are a few color subspace transformation methods~\cite{de2013robust,wang2017algorithmic} which utilize all skin pixels for rPPG measurement. Based on the prior knowledge from traditional methods, a few learning based approaches~\cite{ hsu2017deep,qiu2018evm,niu2018synrhythm,niu2019rhythmnet} are designed as non-end-to-end fashions. ROI based preprocessed signal representations (e.g., time-frequency map~\cite{ hsu2017deep} and spatio-temporal map~\cite{niu2018synrhythm,niu2019rhythmnet}) are generated first, and then learnable models could capture rPPG features from these maps. However, these methods need the strict preprocessing procedure and neglect the global contextual clues outside the pre-defined ROIs. Meanwhile, more and more end-to-end deep learning based rPPG methods~\cite{vspetlik2018visual,chen2018deepphys,yu2019remote1,yu2019remote2,liu2020multi} are developed, which treat facial video frames as input and predict rPPG and other physiological signals directly. However, pure end-to-end methods are easily influenced by the complex scenarios (e.g., with head movement and various illumination conditions) and rPPG-unrelated features can not be ruled out in learning, resulting in large performance decrease~\cite{yu2020autohr} in realistic datasets (e.g., VIPL-HR~\cite{niu2019rhythmnet}). 

Recently, due to its excellent long-range attentional modeling capacities in solving sequence-to-sequence issues, transformer~\cite{lin2021survey,han2020survey} has been successfully applied in many artificial intelligence tasks such as natural language processing (NLP)~\cite{vaswani2017attention}, image~\cite{dosovitskiy2020image} and video~\cite{bertasius2021space} analysis. Similarly, rPPG measurement from facial videos can be treated as a video sequence to signal sequence problem, where the long-range contextual clues should be exploited for semantic modeling. As shown in Fig.~\ref{fig:Figure1}, rPPG clues from different skin regions and temporal locations (e.g., signal trajectories around t1, t2, and t3) share similar properties (e.g., trends with rising edge first then falling edge later and relative high magnitudes), which can be utilized for long-range feature modeling and enhancement. However, different from the most video tasks aiming at huge motion representation, facial rPPG measurement focuses on capturing subtle skin color changes, which makes it challenging for global spatio-temporal perception. Furthermore, video-based rPPG measurement is usually a long-time monitoring task, and it is challenging to design and train transformers with long video sequence inputs.

Motivated by the discussions above, we propose an end-to-end video transformer architecture, namely PhysFormer, for remote physiological measurement. On one hand, the cascaded temporal difference transfomer blocks in PhysFormer benefit the rPPG feature enhancement via global spatio-temporal attention based on the fine-grained temporal skin color differences. On the other hand, to alleviate the interference-induced overfitting issue and complement the weak temporal supervision signals, elaborate supervision in frequency domain is designed, which helps PhysFormer learn more intrinsic rPPG-aware features. 

The contributions of this work are as follows:

\vspace{-0.5em}
\begin{itemize}
\setlength\itemsep{-0.1em}
\vspace{-0.5em}
    
    \item We propose the PhysFormer, which mainly consists of a powerful video temporal difference transformer backbone. To our best knowledge, it is the first time to explore the long-range spatio-temporal relationship for reliable rPPG measurement. 

    \item We propose an elaborate recipe to supervise PhysFormer with label distribution learning and curriculum learning guided dynamic loss in frequency domain to learn efficiently and alleviate overfitting.

    \item We conduct intra- and cross-dataset testings and show that the proposed PhysFormer achieves superior or on par state-of-the-art performance without pretraining on large-scale datasets like ImageNet-21K.
    
\end{itemize}

\section{Related Work}


\noindent\textbf{Remote physiological measurement.}\quad      
An early study of rPPG-based physiological measurement was reported in~\cite{verkruysse2008remote}. Plenty of traditional hand-crafted approaches have been developed on this field since then. Selective merging information from different color channels~\cite{poh2010non,poh2010advancements,li2014remote} or different ROIs~\cite{lam2015robust,li2014remote} are proven to be efficient for subtle rPPG signal recovery. To improve the signal-to-noise-ratio of the recovered rPPG signals, several signal decomposition methods such as independent component analysis (ICA)~\cite{poh2010non,poh2010advancements,lam2015robust} and matrix completion~\cite{tulyakov2016self} are also proposed. 
In recent years, deep learning based approaches dominate the field of rPPG measurement due to the strong spatio-temporal representation capabilities. On one hand, facial ROI based spatial-temporal signal maps~\cite{niu2018synrhythm,niu2019robust,lu2021hr,niu2020video,lu2021dual} are developed, which alleviate the interference from non-skin regions. Based on these signal maps, 2D-CNNs are utilized for rPPG feature extraction. On the other hand, end-to-end spatial networks~\cite{vspetlik2018visual,chen2018deepphys} and spatio-temporal models~\cite{yu2019remote1,yu2019remote2,yu2020autohr,liu2020multi,liu2021efficientphys,nowara2021benefit,gideon2021way} are developed, which could recover rPPG signals from the facial video directly. However, previous methods only consider the spatio-temporal rPPG features from adjacent frames and neglect the long-range relationship among quasi-periodic rPPG features.

\begin{figure*}
\centering
\includegraphics[scale=0.46]{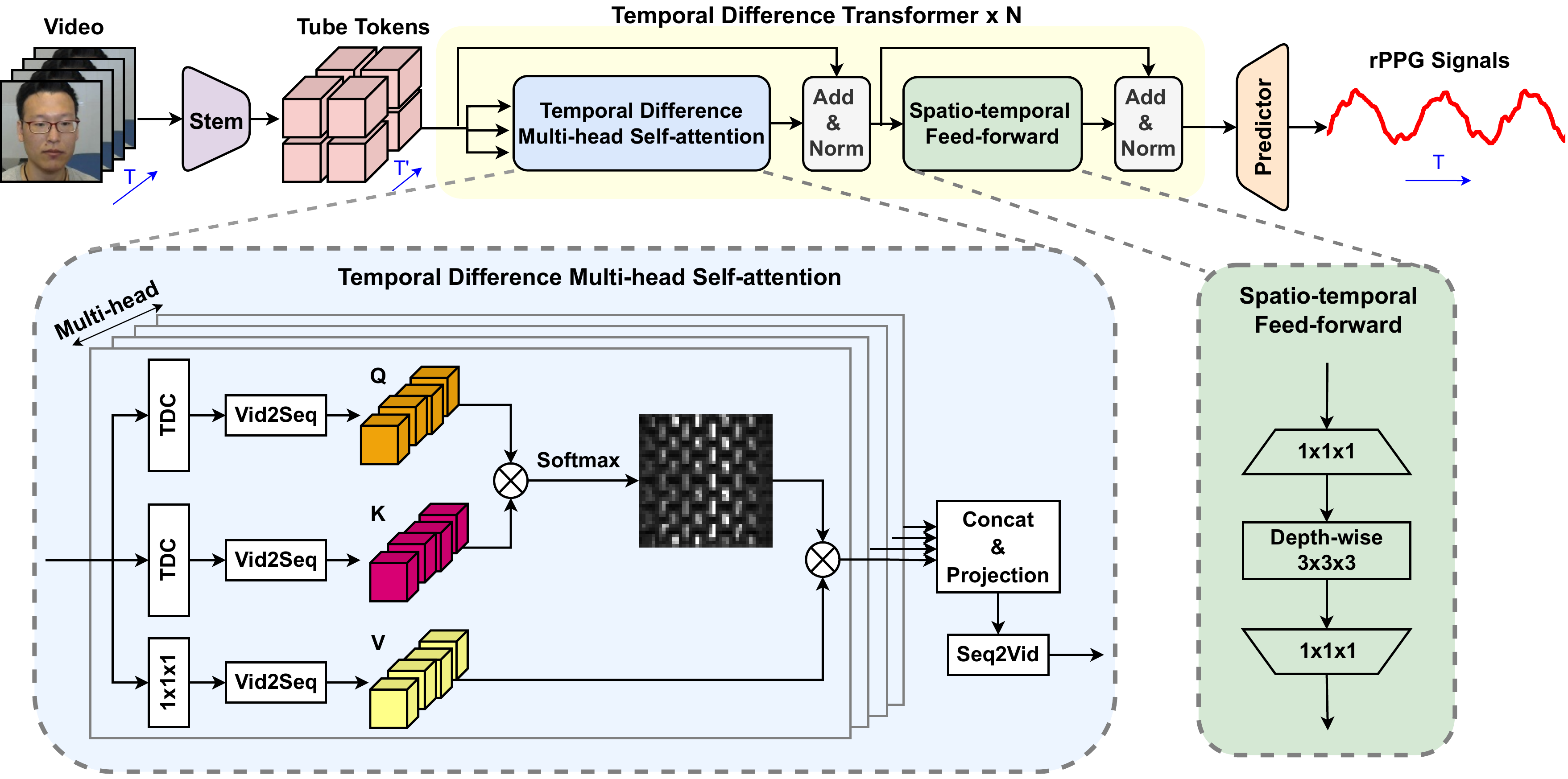}
\vspace{-0.8em}
  \caption{\small{
  Framework of the PhysFormer. It consists of a shallow stem, a tube tokenizer, several temporal difference transformers, and a rPPG predictor head. The temporal difference transformer is formed from the Temporal Difference Multi-head Self-attention (TD-MHSA) and Spatio-temporal Feed-forward (ST-FF) modules, which enhances the global and local spatio-temporal representation, respectively. `TDC' is short for the temporal difference convolution~\cite{yu2020autohr,yu2021searching}.
  }
  }
\label{fig:PhysFormer}
\vspace{-1.0em}
\end{figure*}

\vspace{0.5em}
\noindent\textbf{Transformer for vision tasks.}\quad  
Transformer~\cite{lin2021survey} is proposed in~\cite{vaswani2017attention} to model sequential data in the field of NLP. Then vision transformer (ViT)~\cite{dosovitskiy2020image} is proposed recently by feeding transformer with sequences of image patches for image classification. Many other ViT variants ~\cite{han2020survey,khan2021transformers,touvron2021training,liu2021swin,yuan2021tokens,wang2021pyramid,han2021transformer,chen2021crossvit,ding2021hr} are proposed from then, which achieve promising performance compared with its counterpart CNNs for image analysis tasks~\cite{carion2020end,zheng2021rethinking,he2021transreid}. 
Recently, some works introduce vision transformer for video understanding tasks such as action recognition~\cite{arnab2021vivit,fan2021multiscale,neimark2021video,girdhar2019video,liu2021video,bulat2021space,bertasius2021space}, action detection~\cite{zhao2021tuber,liu2021end,wang2021temporal,xu2021long}, video super-resolution~\cite{cao2021video}, video inpainting~\cite{zeng2020learning,liu2021fuseformer}, and 3D animation~\cite{chen2021geometry,chen2021aniformer}. Some works~\cite{neimark2021video,girdhar2019video} conduct temporal contextual modeling with transformer based on single-frame features from pretrained 2D networks, while other works~\cite{bertasius2021space,arnab2021vivit,liu2021video,bulat2021space,fan2021multiscale} mine the spatio-temporal attentions via video transformer directly. Most of these works are incompatible for long-video-sequence ($\textgreater$150 frames) signal regression task. There are two related works~\cite{yu2021transrppg,liu2021efficientphys} using ViT for rPPG feature representation. TransRPPG~\cite{yu2021transrppg} extracts rPPG features from the preprocessed signal maps via ViT for face 3D mask presentation attack detection~\cite{yu2021deep}. Based on the temporal shift networks~\cite{liu2020multi,lin2019tsm}, EfficientPhys-T~\cite{liu2021efficientphys} adds several swin transformer~\cite{liu2021swin} layers for global spatial attention. Different from these two works, the proposed PhysFormer is an end-to-end video transformer, which is able to capture long-range spatio-temporal attentional rPPG features from facial video directly.

\section{Methodology}
\label{sec:method}

We will first introduce the architecture of PhysFormer in Sec.~\ref{sec:PhysFormer}, then introduce label distribution learning for rPPG measurement in Sec.~\ref{sec:distribution}, and at last present the curriculum learning guided dynamic supervision in Sec.~\ref{sec:dynamic}.

\subsection{PhysFormer}
\label{sec:PhysFormer}

As illustrated in Fig.~\ref{fig:PhysFormer}, PhysFormer consists of a shallow stem $\mathbf{E}_{\text{stem}}$, a tube tokenizer $\mathbf{E}_{\text{tube}}$, $N$ temporal difference transformer blocks $\mathbf{E}^{i}_{\text{trans}}$ ($ i=1,...,N$) and a rPPG predictor head. Inspired by the study in~\cite{xiao2021early}, we adopt a shallow stem
to extract coarse local spatio-temporal features, which benefits the fast convergence and clearer subsequent global self-attention. Specifically, the stem is formed by three convolutional blocks with kernel size (1x5x5), (3x3x3) and (3x3x3), respectively. Each convolution operator is cascaded with a batch normalization (BN), ReLU and MaxPool. The pooling layer only halves the spatial dimension. Therefore, given an RGB facial video input $X\in \mathbb{R}^{3\times T\times H\times W}$, the stem output $X_{\text{stem}}=\mathbf{E}_{\text{stem}}(X)$, where $X_{\text{stem}}\in \mathbb{R}^{D\times T\times H/8\times W/8}$, and $D$, $T$, $W$, $H$ indicate channel, sequence length, width, height, respectively. Then $X_{\text{stem}}$ would be partitioned into spatio-temporal tube tokens $X_{\text{tube}}\in \mathbb{R}^{D\times T'\times H'\times W'}$ via the tube tokenizer $\mathbf{E}_{\text{tube}}$. Subsequently, the tube tokens will be forwarded with $N$ temporal difference transformer blocks and obtain the global-local refined rPPG features $X_{\text{trans}}$, which has the same dimensions with $X_{\text{tube}}$. Finally, the rPPG predictor head temporally upsamples, spatially averages, and projects the features $X_{\text{trans}}$ to 1D signal $Y\in \mathbb{R}^{T}$.

\vspace{0.3em}
\noindent\textbf{Tube tokenization.}\quad  
 Here the coarse feature $X_{\text{stem}}$ would be partitioned into non-overlapping tube tokens via $\mathbf{E}_{\text{tube}}(X_{\text{stem}})$, which aggregates the spatio-temporal neighbor semantics and reduces computational costs for the subsequent transformers. Specifically, with the targeted tube size $T_{s}\times H_{s}\times W_{s}$ (the same as the partition step size in non-overlapping setting), the tube token map $X_{\text{tube}}\in \mathbb{R}^{D\times T'\times H'\times W'}$ has length, height and width
\begin{equation} 
T'=\left \lfloor \frac{T}{T_{s}} \right \rfloor, H'=\left \lfloor \frac{H/8}{H_{s}} \right \rfloor, W'=\left \lfloor \frac{W/8}{W_{s}} \right \rfloor.
\label{eq:token}
\end{equation}
Please note that there are no position embeddings after the tube tokenization as the stem at early stage already captures relative spatio-temporal positions.  

\vspace{0.4em}
\noindent\textbf{Temporal difference multi-head self-attention.}\quad  
In self-attention mechanism~\cite{vaswani2017attention,dosovitskiy2020image}, the relationship between the tokens is modeled by the similarity between the projected query-key pairs, yielding the attention score. Instead of point-wise linear projection, we utilize temporal difference convolution (TDC)~\cite{yu2020autohr,yu2021searching} for query ($Q$) and key ($K$) projection, which could capture fine-grained local temporal difference features for subtle color change description. 
TDC with learnable $w$ can be formulated as
\begin{equation} \footnotesize
\begin{split}
\mathrm{TDC}(x)
&=\underbrace{\sum_{p_n\in \mathcal{R}}w(p_n)\cdot x(p_0+p_n)}_{\text{vanilla 3D convolution}}+\theta\cdot (\underbrace{-x(p_0)\cdot\sum_{p_n\in \mathcal{R'}}w(p_n))}_{\text{temporal difference term}}, \\
\label{eq:CDC-T}
\end{split}
\end{equation}
where $p_0$, $\mathcal{R}$ and $\mathcal{R'}$ indicate the current spatio-temporal location, sampled local (3x3x3) neighborhood and sampled adjacent neighborhood, respectively. Then query and key are projected as 
\vspace{-0.3em}
\begin{equation} 
Q = \mathrm{BN}(\mathrm{TDC}(X_{\text{tube}})), K= \mathrm{BN}(\mathrm{TDC}(X_{\text{tube}})). 
\vspace{-0.3em}
\end{equation}
For the value ($V$) projection, point-wise linear projection without BN is utilized. Then $Q,K,V\in \mathbb{R}^{D\times T'\times H'\times W'}$ are flattened into sequence, and separated into $h$ heads ($D_h=D/h$ for each head). For the $i$-th head ($i\leq h$), the self-attention (SA) can be formulated
\vspace{-0.3em}
\begin{equation}
\mathrm{SA}_{i}=\mathrm{Softmax}(Q_{i}K^{T}_{i}/\tau)V_{i},
\vspace{-0.3em}
\end{equation}
where $\tau$ controls the sparsity. We find that the default setting $\tau=\sqrt{D_h}$ in~\cite{vaswani2017attention,dosovitskiy2020image} performs poorly for rPPG measurement. According to the periodicity of rPPG features, we use smaller $\tau$ value to obtain sparser attention activation. The corresponding study can be found in Table~\ref{tab:ablation2}. The output of TD-MHSA is the concatenation of SA from all heads and then with a linear projection $U\in \mathbb{R}^{D\times D}$
\vspace{-0.3em}
\begin{equation} 
\text{TD-MHSA} = \mathrm{Concat}(\mathrm{SA}_{1}; \mathrm{SA}_{2};...; \mathrm{SA}_{h})U.
\vspace{-0.3em}
\end{equation}
As illustrated in Fig.~\ref{fig:PhysFormer}, residual connection and layer normalization (LN)  would be conducted after TD-MHSA.

\vspace{0.5em}
\noindent\textbf{Spatio-temporal feed-forward.}
The vanilla feed-forward network consists of two linear transformation layers, where the hidden dimension $D'$ between two layers is expanded to learn a richer feature representation. In contrast, we introduce a depthwise 3D convolution (with BN and nonlinear activation) between these two layers with extra slight computational cost but remarkable performance improvement. The benefits are two-fold: 1) as a complementation of TD-MHSA, ST-FF could refine the local inconsistency and parts of noisy features; 2) richer locality provides TD-MHSA sufficient relative position cues.

\subsection{Label Distribution Learning}
\label{sec:distribution}
Similar to the facial age estimation task~\cite{geng2013facial,gao2018age} that faces at close ages look quite similar, facial rPPG signals with close HR values usually have similar periodicity. Inspired by this observation, instead of considering each facial video as an instance with one label (HR), we regard each facial video as an instance associated with a label distribution. The label distribution covers a certain number of class labels, representing the degree that each label describes the instance. Through this way, one facial video can contribute to both targeted HR value and its adjacent HRs.

To consider the similarity information among HR classes during the training stage, we model the rPPG-based HR estimation problem as a specific $L$-class multi-label classification problem, where $L$=139 in our case (each integer HR value within [42, 180] bpm as a class). A label distribution $\mathbf{p}= \left\{ p_1,p_2,...,p_L\right\}\in \mathbb{R}^L$ is assigned to each facial video $X$. It is assumed that each entry of $\mathbf{p}$ is a real value in the range [0,1] such that $\sum_{k=1}^{L}p_k=1$. We consider the Gaussian distribution function, centred at the ground truth HR label $Y_{\text{HR}}$ with the standard deviation $\sigma$, to construct the corresponding label distribution $\mathbf{p}$. 
\begin{equation} 
p_k=\frac{1}{\sqrt{2\pi}\sigma}\exp\left ( -\frac{(k-(Y_{HR}-41))^2}{2\sigma^2} \right ).
\end{equation}
The label distribution loss can be formulated as $\mathcal{L}_{\text{LD}}=\mathrm{KL}(\mathbf{p}, \mathrm{Softmax}(\mathbf{\hat{p}}))$, where divergence measure $\mathrm{KL}(\cdot)$ denotes the Kullback-Leibler (KL) divergence~\cite{gao2017deep}, and $\mathbf{\hat{p}}$ is the power spectral density (PSD) of predicted rPPG signals. 


Please note that the previous work~\cite{niu2017continuous} also considers the distribution learning for HR estimation. However, it is totally different with our work: 1) the motivation in~\cite{niu2017continuous} is to smooth the temporal HR outliers caused by facial movements across continuous video clips, while our work is more generic, aiming at efficient feature learning across adjacent labels under limited-scale training data; 2) the technique used in~\cite{niu2017continuous} is after a post-HR-estimation for the handcrafted rPPG signals, while our work is to design a reasonable supervision signal $\mathcal{L}_{\text{LD}}$ for PhysFormer.

\subsection{Curriculum Learning Guided Dynamic Loss}
\label{sec:dynamic}

Curriculum learning~\cite{bengio2009curriculum}, as a major machine learning regime with philosophy of easy-to-hard curriculum, is utilized to train PhysFormer. In the rPPG measurement task, the supervision signals from temporal domain (e.g., mean square error loss~\cite{chen2018deepphys}, negative Pearson loss~\cite{yu2019remote1,yu2019remote2}) and frequency domain (e.g., cross-entropy loss~\cite{niu2020video,yu2020autohr}, signal-to-noise ratio loss~\cite{vspetlik2018visual}) provide different extents of constraints for model learning. The former one gives signal-trend-level constraints, which is straightforward and easy for model convergence but overfitting after that. In contrast, the latter one with strong constraints on frequency domain enforces the model learning periodic features within target frequency bands, which is hard to converged well due to the realistic rPPG-irrelevant noise. Inspired by the curriculum learning, we propose the dynamic supervision to gradually enlarge the frequency constraints, which alleviates the overfitting issue and benefits the intrinsic rPPG-aware feature learning gradually. Specifically, exponential increment strategy is adopted, and comparison with other dynamic strategies (e.g., linear increment) will be shown in Table~\ref{tab:ablation3}. The dynamic loss $\mathcal{L}_{\text{overall}}$ can be formulated as
\vspace{-0.1em}
\begin{equation}
\begin{split}
\mathcal{L}_{\text{overall}}&=\underbrace{\alpha\cdot\mathcal{L}_{\text{time}}}_{\text{temporal}}+\underbrace{\beta\cdot(\mathcal{L}_{\text{CE}}+\mathcal{L}_{\text{LD}})}_{\text{frequency}},\\
\beta&=\beta_{0}\cdot(\eta^{({\text{Epoch}}_{\text{current}}-1)/{\text{Epoch}}_{\text{total}}}),
\vspace{-0.1em}
\end{split}
\end{equation}
where hyperparameters $\alpha$, $\beta_{0}$ and $\eta$ equal to 0.1, 1.0 and 5.0, respectively. Negative Pearson loss~\cite{yu2019remote1,yu2019remote2} and frequency cross-entropy loss~\cite{niu2020video,yu2020autohr} are adopted as $\mathcal{L}_{\text{time}}$ and $\mathcal{L}_{\text{CE}}$, respectively. With the dynamic supervision, PhysFormer could  perceive better signal trend at the beginning while such perfect warming up facilitates the gradually stronger frequency knowledge learning later.

\section{Experimental Evaluation}
\label{sec:experiemnts}
Experiments of rPPG-based physiological measurement for three types of physiological signals, i.e., heart rate (HR), heart rate variability (HRV), and respiration frequency (RF), are conducted on four benchmark datasets (VIPL-HR~\cite{niu2019rhythmnet}, MAHNOB-HCI~\cite{soleymani2011multimodal}, MMSE-HR~\cite{tulyakov2016self}, and OBF~\cite{li2018obf}).

\subsection{Datasets and Performance Metrics}

\label{sec:dataset} \textbf{VIPL-HR}~\cite{niu2019rhythmnet} is a large-scale dataset for remote physiological measurement under less-constrained scenarios. It contains 2,378 RGB videos of 107 subjects recorded with different head movements, lighting conditions and acquisition devices. \textbf{MAHNOB-HCI}~\cite{soleymani2011multimodal} is one of the most widely used benchmark for remote HR measurement evaluations. It includes 527 facial videos of with 61 fps framerate and 780x580 resolution from 27 subjects. \textbf{MMSE-HR}~\cite{tulyakov2016self} is a dataset including 102 RGB videos from 40 subjects, and the raw resolution of each video is at 1040x1392. \textbf{OBF}~\cite{li2018obf} is a high-quality dataset for remote physiological signal measurement. It contains 200 five-minute-long RGB videos with 60 fps framerate recorded from 100 healthy adults.

Average HR estimation task is evaluated on all four datasets while HRV and RF estimation tasks on high-quality OBF~\cite{li2018obf} dataset. Specifically, we follow existing methods~\cite{yu2019remote2,niu2020video,lu2021dual} and report low frequency (LF), high frequency (HF), and LF/HF ratio for HRV and RF estimation. We report the most commonly used performance metrics for evaluation, including the standard deviation (SD), mean absolute error (MAE), root mean square error (RMSE), and Pearson's correlation coefficient ($r$).

\subsection{Implementation Details}
\label{sec:Details}

Our proposed method is implemented with Pytorch. For each video clip, we use the MTCNN face detector~\cite{zhang2016joint} to crop the enlarged face area at the first frame and fix the region through the following frames. The videos in MAHNOB-HCI and OBF are downsampled to 30 fps for efficiency. The settings $N$=12, $h$=4, $D$=96, $D'$=144 are used for PhysFormer while $\theta$=0.7 and $\tau$=2.0 for TD-MHSA. The targeted tube size $T_{s}\times H_{s}\times W_{s}$ equals to 4$\times$4$\times$4. In the training stage, we randomly sample RGB face clips with size 160$\times$128$\times$128 ($T\times H\times W$) as model inputs. Random horizontal flipping and temporally up/down-sampling~\cite{yu2020autohr} are used for data augmentation. The PhysFormer is trained with Adam optimizer and the initial learning rate and weight decay are 1e-4 and 5e-5, respectively. We cannot find obvious performance improvement using AdamW optimizer. We train models with 25 epochs with fixed setting $\alpha$=0.1 for temporal loss while exponentially increased parameter $\beta \in [1,5]$ for frequency losses. We set $\sigma$=1.0 for label distribution learning. The batch size is 4 on one 32G V100 GPU. In the testing stage, similar to~\cite{niu2019rhythmnet}, we uniformly separate 30-second videos into three short clips with 10 seconds, and then video-level HR is calculated via averaging HRs from three short clips.

\begin{table}[t]\small
\centering
\caption{Intra-dataset testing results on VIPL-HR~\cite{niu2019rhythmnet}. The symbols $\blacktriangle$, $\blacklozenge$ and $\star$ denote traditional, non-end-to-end learning based and end-to-end learning based methods, respectively. Best results are marked in \textbf{bold} and second best in \underline{underline}.} \label{tab:ResultsVIPL}

\vspace{-0.8em}

\resizebox{0.47\textwidth}{!} {\begin{tabular}{l c c c c} 
 \toprule[1pt]
 Method & \begin{tabular}[c]{@{}c@{}}SD $\downarrow$ \\(bpm)\end{tabular} & \begin{tabular}[c]{@{}c@{}}MAE $\downarrow$ \\(bpm)\end{tabular}  & \begin{tabular}[c]{@{}c@{}}RMSE $\downarrow$ \\(bpm)\end{tabular}  & $r$ $\uparrow$\\
 \midrule
 Tulyakov2016~\cite{tulyakov2016self}$\blacktriangle$ & 18.0 & 15.9 & 21.0 &  0.11\\
 POS~\cite{wang2017algorithmic}$\blacktriangle$ & 15.3 & 11.5 & 17.2 & 0.30 \\
 CHROM~\cite{de2013robust}$\blacktriangle$ & 15.1 & 11.4 & 16.9 & 0.28 \\ 
 \midrule
 RhythmNet~\cite{niu2019rhythmnet}$\blacklozenge$ & 8.11 & 5.30 & 8.14 &  0.76\\
 ST-Attention~\cite{niu2019robust}$\blacklozenge$ & 7.99 & 5.40 & 7.99 &  0.66\\
 NAS-HR~\cite{lu2021hr}$\blacklozenge$ & 8.10 & 5.12 & 8.01 & \underline{0.79}\\
 CVD~\cite{niu2020video}$\blacklozenge$ & 7.92 & 5.02 &7.97 &  \underline{ 0.79}\\
  Dual-GAN~\cite{lu2021dual}$\blacklozenge$ & \textbf{7.63} & \textbf{4.93} & \textbf{7.68} &  \textbf{0.81}\\
 \midrule
 I3D~\cite{carreira2017quo}$\star$ & 15.9  & 12.0 & 15.9 & 0.07\\
  PhysNet~\cite{yu2019remote1}$\star$ & 14.9 & 10.8 & 14.8 & 0.20 \\
 DeepPhys~\cite{chen2018deepphys}$\star$ & 13.6 & 11.0  & 13.8 & 0.11 \\ 

AutoHR~\cite{yu2020autohr}$\star$ & 8.48
 & 5.68 & 8.68 & 0.72\\

 \textbf{PhysFormer (Ours)$\star$} & \underline{7.74
} & \underline{4.97} & \underline{7.79} & 0.78\\

 \bottomrule[1pt]
 \end{tabular}}
 \vspace{-1.0em}
\end{table}

\begin{table*}[t]\footnotesize
\begin{center}
\caption{Performance comparison of HR and RF measurement as well as HRV analysis on OBF~\cite{li2018obf}. }

\vspace{-1.0em}

\label{tab:OBFrPPGNet}
\centering
\begin{tabular}{p{2.6cm} p{0.5cm} p{0.5cm} p{0.6cm} p{0.5cm} p{0.5cm} p{0.6cm}  p{0.5cm} p{0.5cm} p{0.6cm} p{0.5cm} p{0.5cm} p{0.6cm} p{0.5cm} p{0.5cm} p{0.6cm}}

\toprule[1pt]
& \multicolumn{3}{c}{HR(bpm)} &  \multicolumn{3}{c}{RF(Hz)} &  \multicolumn{3}{c}{LF(u.n)} &  \multicolumn{3}{c}{HF(u.n)} &  \multicolumn{3}{c}{LF/HF}\\
  
\cmidrule(lr){2-4} \cmidrule(lr){5-7}
\cmidrule(lr){8-10} \cmidrule(lr){11-13}
\cmidrule(lr){14-16}

  \multicolumn{1}{c}{Method} & \multicolumn{1}{c}{SD} & RMSE & \multicolumn{1}{c}{$r$}  &  \multicolumn{1}{c}{SD} & RMSE & \multicolumn{1}{c}{$r$}  &  \multicolumn{1}{c}{SD} & RMSE & \multicolumn{1}{c}{$r$}  & \multicolumn{1}{c}{SD} & RMSE & \multicolumn{1}{c}{$r$}  &  \multicolumn{1}{c}{SD} & RMSE & \multicolumn{1}{c}{$r$}\\

 \midrule
 ROI\_green~\cite{li2018obf}$\blacktriangle$   & 2.159 & 2.162 & 0.99   & 0.078 & 0.084  & 0.321  & 0.22  & 0.24 & 0.573 & 0.22  & 0.24 & 0.573 & 0.819  & 0.832 & 0.571\\
 
 CHROM~\cite{de2013robust}$\blacktriangle$   & 2.73 & 2.733 & 0.98   & 0.081 & 0.081  & 0.224  & 0.199  & 0.206 & 0.524 & 0.199  & 0.206 & 0.524 & 0.83  & 0.863 & 0.459\\
 
 POS~\cite{wang2017algorithmic}$\blacktriangle$   & 1.899 & 1.906 & 0.991   & 0.07 & 0.07  & 0.44  & 0.155  & 0.158 & 0.727 & 0.155  & 0.158 & 0.727 & 0.663  & 0.679 & 0.687\\
  \midrule

 CVD~\cite{niu2020video}$\blacklozenge$  & \underline{1.257} & \underline{1.26} & \underline{0.996}  & \underline{0.058} & \underline{0.058}  & \underline{0.606} & \underline{0.09}  & \underline{0.09} & \textbf{0.914} & \underline{0.09}  & \underline{0.09} & \textbf{0.914} & \underline{0.453}  & \underline{0.453} & \underline{0.877}\\

 \midrule

 rPPGNet~\cite{yu2019remote2}$\star$  & 1.756 & 1.8 & 0.992  & 0.064 & 0.064  & 0.53 & 0.133  & 0.135 & 0.804 & 0.133  & 0.135 & 0.804 & 0.58  & 0.589 & 0.773\\
 
  \textbf{PhysFormer (Ours)}$\star$  & \textbf{0.804} & \textbf{0.804} & \textbf{0.998}  & \textbf{0.054} & \textbf{0.054}  & \textbf{0.661} & \textbf{0.085}  & \textbf{0.086} & \underline{0.912} &  \textbf{0.085}  & \textbf{0.086} & \underline{0.912} & \textbf{0.389}  & \textbf{0.39} & \textbf{0.896}\\
 
\bottomrule[1pt]
\end{tabular}
\end{center}
\vspace{-2.2em}
\end{table*}

\begin{table}[t]\small
\centering
\caption{Intra-dataset results on MAHNOB-HCI~\cite{soleymani2011multimodal}.} \label{tab:ResultsMAHNOB}

\vspace{-1.0em}

\resizebox{0.45\textwidth}{!} {\begin{tabular}{l c c c c} 
 \toprule[1pt]
 Method & \begin{tabular}[c]{@{}c@{}}SD $\downarrow$ \\(bpm)\end{tabular} & \begin{tabular}[c]{@{}c@{}}MAE $\downarrow$ \\(bpm)\end{tabular}  & \begin{tabular}[c]{@{}c@{}}RMSE $\downarrow$ \\(bpm)\end{tabular}  & $r$ $\uparrow$\\
 \midrule
 Poh2011~\cite{poh2010advancements}$\blacktriangle$ & 13.5 & - & 13.6 & 0.36 \\ 
 CHROM~\cite{de2013robust}$\blacktriangle$ & - & 13.49 & 22.36 & 0.21 \\
 Li2014~\cite{li2014remote}$\blacktriangle$ & 6.88 & - & 7.62 & 0.81\\
 Tulyakov2016~\cite{tulyakov2016self}$\blacktriangle$ & 5.81 & 4.96 & 6.23 & 0.83\\
  \midrule
 SynRhythm~\cite{niu2018synrhythm}$\blacklozenge$ & 10.88 & - & 11.08 & - \\ 
 RhythmNet~\cite{niu2019rhythmnet}$\blacklozenge$ & \underline{3.99} & - & 3.99  & \textbf{0.87} \\ 
 \midrule
 HR-CNN~\cite{vspetlik2018visual}$\star$ & - & 7.25 & 9.24 & 0.51 \\
  rPPGNet~\cite{yu2019remote2}$\star$ & 7.82 & 5.51 & 7.82 & 0.78 \\
  DeepPhys~\cite{chen2018deepphys}$\star$ & - & 4.57 & - & -\\
 AutoHR~\cite{yu2020autohr}$\star$ & 4.73 & 3.78 & 5.10 & \underline{0.86}\\
 Meta-rPPG~\cite{lee2020meta}$\star$ & 4.9 & \textbf{3.01} & \textbf{3.68} & 0.85\\
 
 \textbf{PhysFormer (Ours)$\star$} & \textbf{3.87} & \underline{3.25} & \underline{3.97} & \textbf{0.87}\\
 
 \bottomrule[1pt]
 \end{tabular}}
 \vspace{-0.8em}
\end{table}

\subsection{Intra-dataset Testing}

\noindent\textbf{HR estimation on VIPL-HR.} \quad   In these experiments, we follow~\cite{niu2019rhythmnet} and use a
subject-exclusive 5-fold cross-validation protocol on VIPL-HR. As shown in Table~\ref{tab:ResultsVIPL}, all three traditional methods (Tulyakov2016~\cite{tulyakov2016self}, POS~\cite{wang2017algorithmic} and CHROM~\cite{de2013robust}) perform poorly due to the complex scenarios (e.g., large head movement and various illumination) in the VIPL-HR dataset. Similarly, the existing end-to-end learning based methods (e.g., PhysNet~\cite{yu2019remote1}, DeepPhys~\cite{chen2018deepphys}, and AutoHR~\cite{yu2020autohr}) predict unreliable HR values with large RMSE compared with non-end-to-end learning approaches (e.g., RhythmNet~\cite{niu2019rhythmnet}, ST-Attention~\cite{niu2019robust}, NAS-HR~\cite{lu2021hr}, CVD~\cite{niu2020video}, and Dual-GAN~\cite{lu2021dual}). Such the large performance margin might be caused by the coarse and overfitted rPPG features extracted from the end-to-end models. In contrast, all five non-end-to-end methods first extract fine-grained signal maps from multiple facial ROIs, and then more dedicated rPPG clues would be extracted via the cascaded models. Without strict and heavy preprocessing procedure in~\cite{niu2019rhythmnet,niu2019robust,lu2021hr,niu2020video,lu2021dual}, our proposed PhysFormer can be trained from scratch on facial videos directly, and achieves comparable performance with state-of-the-art non-end-to-end learning based method Dual-GAN~\cite{lu2021dual}. It indicates that PhysFormer is able to learn the intrinsic and periodic rPPG-aware features automatically.

\noindent\textbf{HR estimation on MAHNOB-HCI.} \quad   
For the HR estimation tasks on MAHNOB-HCI, similar to~\cite{yu2019remote2}, subject-independent 9-fold cross-validation protocol is adopted. In consideration of the convergence difficulty due to the low illumination and high compression videos in MAHNOB-HCI, we finetune the VIPL-HR pretrained model on MAHNOB-HCI for further 15 epochs. The HR estimation results are shown in Table~\ref{tab:ResultsMAHNOB}. The proposed PhysFormer achieves the lowest SD (3.87 bpm) and highest $r$ (0.87) among the traditional, non-end-to-end learning, and end-to-end learning methods, which indicates the reliability of the learned rPPG features from PhysFormer under sufficient supervision. Our performance is on par with the latest end-to-end learning method Meta-rPPG~\cite{lee2020meta} without transductive adaptation from target frames.

\begin{table}[t]\small
\centering
\caption{Cross-dataset results on MMSE-HR~\cite{tulyakov2016self}.} \label{tab:ResultsMMSE}

\vspace{-1.0em}

\resizebox{0.46\textwidth}{!} {\begin{tabular}{l c c c c} 
 \toprule[1pt]
 Method & \begin{tabular}[c]{@{}c@{}}SD $\downarrow$ \\(bpm)\end{tabular}& \begin{tabular}[c]{@{}c@{}}MAE $\downarrow$ \\(bpm)\end{tabular} & \begin{tabular}[c]{@{}c@{}}RMSE $\downarrow$ \\(bpm)\end{tabular}  & $r$ $\uparrow$\\
 \midrule
 Li2014~\cite{li2014remote}$\blacktriangle$ & 20.02 & - & 19.95 & 0.38\\
 CHROM~\cite{de2013robust}$\blacktriangle$ & 14.08 & - & 13.97 & 0.55 \\
 Tulyakov2016~\cite{tulyakov2016self}$\blacktriangle$ & 12.24 & - & 11.37 & 0.71\\
 \midrule
 ST-Attention~\cite{niu2019robust}$\blacklozenge$ & 9.66 & - & 10.10 & 0.64 \\ 
 RhythmNet~\cite{niu2019rhythmnet}$\blacklozenge$ & 6.98 & - & 7.33 & 0.78 \\ 
  CVD~\cite{niu2020video}$\blacklozenge$ & 6.06  & - &  6.04 & 0.84 \\ 
 \midrule
 PhysNet~\cite{yu2019remote1}$\star$ & 12.76 & - & 13.25 & 0.44 \\
 TS-CAN~\cite{liu2020multi}$\star$ & - & 3.85 & 7.21 & 0.86 \\
AutoHR~\cite{yu2020autohr}$\star$ & \underline{5.71} & - & 5.87 & \underline{0.89}\\
EfficientPhys-C~\cite{liu2021efficientphys}$\star$ & - &  \underline{2.91} & \underline{5.43} & \textbf{0.92}\\
EfficientPhys-T1~\cite{liu2021efficientphys}$\star$ & - & 3.48 & 7.21 & 0.86\\
\textbf{PhysFormer (Ours)$\star$} & \textbf{5.22} & \textbf{2.84} & \textbf{5.36} & \textbf{0.92}\\

 \bottomrule[1pt]
 \end{tabular}}
 \vspace{-0.8em}
\end{table}

\vspace{0.3em}
\noindent\textbf{HR, HRV and RF estimation on OBF.} \quad  We also conduct experiments for three types of physiological signals, i.e., HR, RF, and HRV measurement on the OBF~\cite{li2018obf} dataset. Following~\cite{yu2019remote2,niu2020video}, we use a 10-fold subject-exclusive protocol for all experiments. All the results are shown in Table~\ref{tab:OBFrPPGNet}. From the results, we can see that the proposed approach outperforms the existing state-of-the-art traditional (ROI\_green~\cite{li2018obf}, CHROM~\cite{de2013robust}, POS~\cite{wang2017algorithmic}) and end-to-end learning (rPPGNet~\cite{yu2019remote2}) methods by a large margin on all evaluation metrics for HR, RF and all HRV features. The proposed PhysFormer also gives more accurate estimation in terms of HR, RF, and LF/HF compared with the preprocessed signal map based non-end-to-end learning method CVD~\cite{niu2020video}. These results indicate that PhysFormer could not only handle the average HR estimation task but also give a promising prediction of the rPPG signal for RF measurement and HRV analysis, which shows its potential in many healthcare applications.


\vspace{-0.3em}
\subsection{Cross-dataset Testing}
\vspace{-0.3em}
Besides of the intra-dataset testings on the VIPL-HR, MAHNOB-HCI, and OBF datasets, we also conduct cross-dataset testing on MMSE-HR~\cite{tulyakov2016self} following the protocol of~\cite{niu2019rhythmnet}. The models trained on VIPL-HR are directly tested on MMSE-HR. All the results of the proposed approach and the state-of-the-art methods are shown in Table~\ref{tab:ResultsMMSE}. It is clear that the proposed PhysFormer generalizes well in unseen domain. It is worth noting that PhysFormer achieves the lowest SD (5.22 bpm), MAE (2.84 bpm), RMSE (5.36 bpm) as well as the highest $r$ (0.92) among the traditional, non-end-to-end learning and end-to-end learning based methods, indicating 1) the predicted HRs are highly correlated with the ground truth HRs, and 2) the model learns domain-invariant intrinsic rPPG-aware features. Compared with the spatio-temporal transformer based EfficientPhys-T1~\cite{liu2021efficientphys}, our proposed PhysFormer is able to predict more accurate physiological signals, which indicates the effectiveness of the long-range spatio-temporal attention.

\begin{table}\small
\centering

\caption{Ablation of Tube Tokenization of PhysFormer. The three dimensions in tensors indicate length$\times$ height$\times$width.}

\vspace{-0.9em}

\scalebox{0.87}{\begin{tabular}{l|c | c |c}
\toprule[1pt]
Inputs & \begin{tabular}[c]{@{}c@{}} [Stem] \\ Feature Size\end{tabular} & \begin{tabular}[c]{@{}c@{}} [Tube Size] \\ Token Numbers\end{tabular} & \begin{tabular}[c]{@{}c@{}}RMSE $\downarrow$ \\(bpm)\end{tabular} \\
\hline
$160 \times 128 \times 128$ & \begin{tabular}[c]{@{}c@{}} [$\bigtimes$] \\ $160 \times 128 \times 128$ \end{tabular} & \begin{tabular}[c]{@{}c@{}} $[4 \times 32 \times 32]$ \\ $40 \times 4 \times 4$ \end{tabular} & 10.62 \\
 \hline
$160 \times 128 \times 128$ & \begin{tabular}[c]{@{}c@{}} [$\surd$] \\ $160 \times 16 \times 16$ \end{tabular} & \begin{tabular}[c]{@{}c@{}} $[4 \times 4 \times 4]$ \\ $40 \times 4 \times 4$ \end{tabular} & \textbf{7.56} \\
 \hline
$160 \times 96 \times 96$ & \begin{tabular}[c]{@{}c@{}} [$\surd$] \\ $160 \times 12 \times 12$ \end{tabular} & \begin{tabular}[c]{@{}c@{}} $[4 \times 4 \times 4]$ \\ $40 \times 3 \times 3$ \end{tabular} & 8.03 \\
 \hline
$160 \times 128 \times 128$ & \begin{tabular}[c]{@{}c@{}} [$\surd$] \\ $160 \times 16 \times 16$ \end{tabular} & \begin{tabular}[c]{@{}c@{}} $[4 \times 16 \times 16]$ \\ $40 \times 1 \times 1$ \end{tabular} & 10.61 \\
 \hline
 $160 \times 128 \times 128$ & \begin{tabular}[c]{@{}c@{}} [$\surd$] \\ $160 \times 16 \times 16$ \end{tabular} & \begin{tabular}[c]{@{}c@{}} $[2 \times 4 \times 4]$ \\ $80 \times 4 \times 4$ \end{tabular} & 7.81 \\
 
\bottomrule[1pt]

\end{tabular}}

\label{tab:tube}
\vspace{-0.5em}
\end{table}

\begin{table}
\centering

\caption{Ablation of TD-MHSA and ST-FF in PhysFormer. }

\vspace{-0.9em}

\scalebox{0.86}{\begin{tabular}{l|c | c |c}
\toprule[1pt]
MHSA & $\tau$ & Feed-forward & RMSE (bpm) $\downarrow$\\
\hline
- & - & ST-FF & 9.81 \\
TD-MHSA & $\sqrt{D_{h}}\approx$ 4.9 & ST-FF & 9.51 \\
TD-MHSA & 2.0 & ST-FF & \textbf{7.56} \\
vanilla MHSA & 2.0 & ST-FF & 10.43 \\
TD-MHSA & 2.0 & vanilla FF & 8.27 \\
 
\bottomrule[1pt]

\end{tabular}}

\label{tab:ablation2}
\vspace{-0.5em}
\end{table}

\begin{table}
\centering

\caption{Ablation of dynamic loss in the frequency domain. The temporal loss $\mathcal{L}_{time}$ is with fixed $\alpha$=0.1 here. `CE' and `LD' denote cross-entropy and label distribution, respectively.}

\vspace{-0.9em}

\scalebox{0.85}{\begin{tabular}{l|c | c |c}
\toprule[1pt]
Frequency loss & $\beta$ & Strategy & RMSE (bpm) $\downarrow$\\
\hline
$\mathcal{L}_{\text{CE}}$ + $\mathcal{L}_{\text{LD}}$  & 1.0 & fixed & 8.48 \\
$\mathcal{L}_{\text{CE}}$ + $\mathcal{L}_{\text{LD}}$  & 5.0 & fixed & 8.86 \\
$\mathcal{L}_{\text{CE}}$ + $\mathcal{L}_{\text{LD}}$  & [1.0, 5.0] & linear & 8.37 \\
$\mathcal{L}_{\text{CE}}$ + $\mathcal{L}_{\text{LD}}$  & [1.0, 5.0] & exponential & \textbf{7.56} \\
$\mathcal{L}_{\text{CE}}$  & [1.0, 5.0] & exponential & 8.09 \\
$\mathcal{L}_{\text{LD}}$  & [1.0, 5.0] & exponential & 8.21 \\
$\mathcal{L}_{\text{LD}}$ (real distribution)  & [1.0, 5.0] & exponential & 8.72 \\
 
\bottomrule[1pt]

\end{tabular}}

\label{tab:ablation3}
\vspace{-1.0em}
\end{table}

\subsection{Ablation Study}
\label{sec:ablation}
We also provide the results of ablation studies for HR estimation on the \textbf{Fold-1} of the VIPL-HR~\cite{niu2019rhythmnet} dataset.

\vspace{0.3em}
\noindent\textbf{Impact of tube tokenization.}\quad   In the default setting of PhysFormer, a shallow stem cascaded with a tube tokenization is used. In this ablation, we consider other four tokenization configurations with or w/o stem. It can be seen from the first row in Table~\ref{tab:tube} that the stem helps the PhysFormer see better~\cite{xiao2021early}, and the RMSE increases dramatically (+3.06 bpm) when w/o the stem. Then we investigate the impacts of the spatial and temporal domains in tube tokenization. It is clear that the result in the fourth row with full spatial projection is quite poor (RMSE=10.61 bpm), indicating the necessity of the spatial attention. In contrast, tokenization with smaller tempos (e.g., [2x4x4]) or spatial inputs (e.g., 160x96x96) reduces performance slightly. 

\vspace{0.3em}
\noindent\textbf{Impact of TD-MHSA and ST-FF.}\quad   
As shown in Table~\ref{tab:ablation2}, both the TD-MHSA and ST-FF play vital roles in PhysFormer. The result in the first row shows that the performance degrades sharply without spatio-temporal attention. Moreover, it can be seen from the last two rows that without TD-MHSA/ST-FF, PhysFormer with vanilla MHSA/FF obtains 10.43/8.27 bpm RMSE. One important finding in this research is that, the temperature $\tau$ influences the MHSA a lot. When the $\tau=\sqrt{D_{h}}$ like previous ViT~\cite{dosovitskiy2020image,arnab2021vivit}, the predicted rPPG signals are unsatisfied (RMSE=9.51 bpm). Regularizing the $\tau$ with smaller value enforces sparser spatio-temporal attention, which is effective for the quasi-periodic rPPG task.

\vspace{0.3em}
\noindent\textbf{Impact of label distribution learning.}\quad 
Besides the temporal loss $\mathcal{L}_{\text{time}}$ and frequency cross-entropy loss $\mathcal{L}_{\text{CE}}$, the ablations w/ and w/o label distribution loss $\mathcal{L}_{\text{LD}}$ are shown in the last four rows of Table~\ref{tab:ablation3}. Although the $\mathcal{L}_{\text{LD}}$ performs slightly worse (+0.12 bpm RMSE) than $\mathcal{L}_{\text{CE}}$, the best performance can be achieved using both losses, indicating the effectiveness of explicit distribution constraints for extreme-frequency interference alleviation and adjacent label knowledgement propagation. It is interesting to find from the last two rows that using real PSD distribution from groundtruth PPG signals as $\mathbf{p}$, the performance is inferior due to the lack of an obvious peak and partial noise. We can also find from the Fig.~\ref{fig:distribution}(a) that the $\sigma$ ranged from 0.9 to 1.2 for $\mathcal{L}_{\text{LD}}$ are suitable to achieve good performance.     


\begin{figure}
\centering
\includegraphics[scale=0.19]{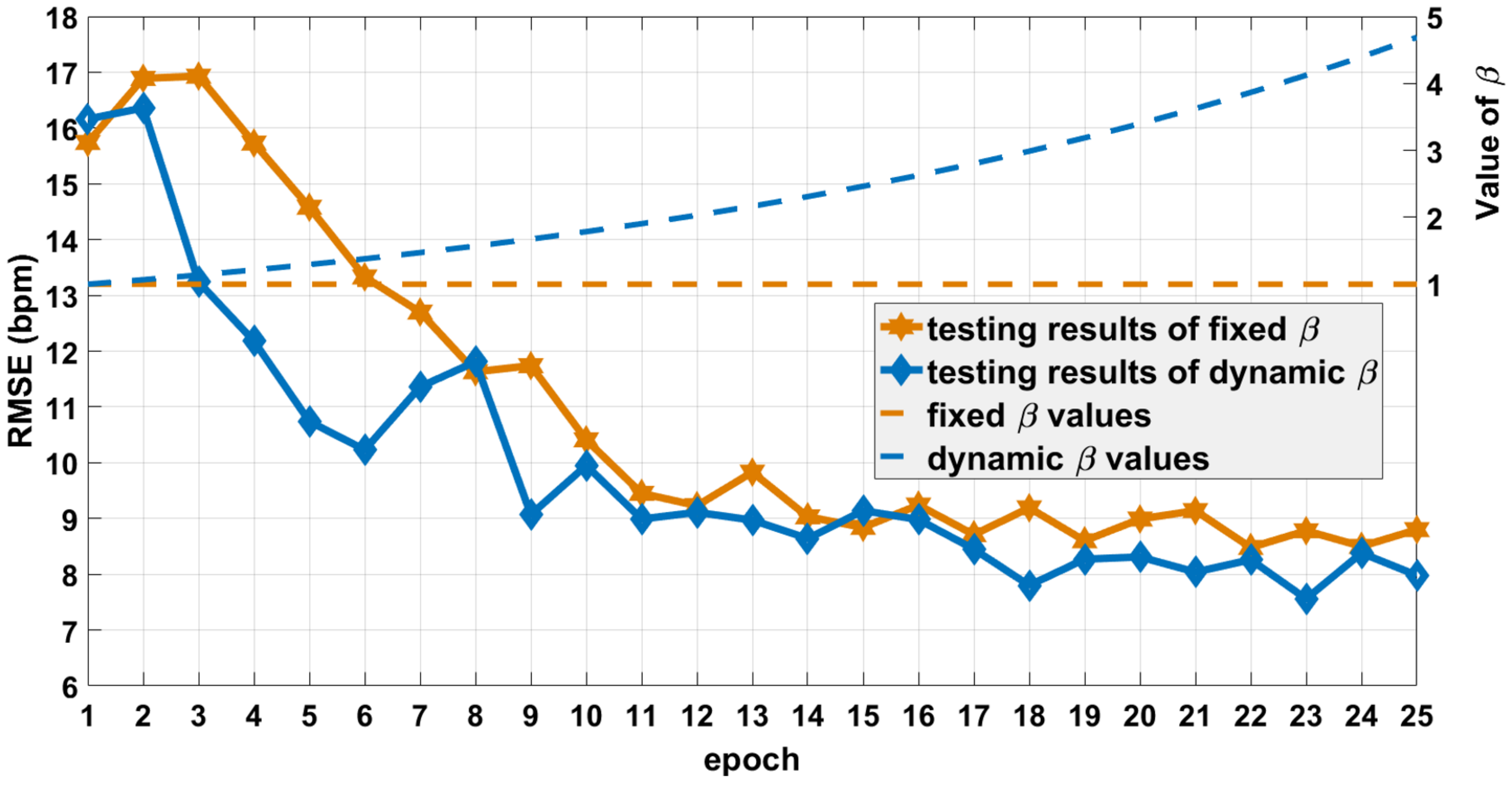}

\vspace{-1.0em}
  \caption{\small{
 Testing results of fixed and dynamic frequency supervisions on the Fold-1 of VIPL-HR. }
  }
\label{fig:dynamicloss}
\vspace{-1.2em}
\end{figure}

\vspace{0.3em}
\noindent\textbf{Impact of dynamic supervision.}\quad 
Fig.~\ref{fig:dynamicloss} illustrates the testing performance on Fold-1 VIPL-HR when training with fixed and dynamic supervision. It is clear that with exponential increased frequency loss, models in the blue curve converge faster and achieve smaller RMSE. We also compare several kinds of fixed and dynamic strategies in Table~\ref{tab:ablation3}. The results in the first four rows indicate 1) using fixed higher $\beta$ leads to poorer performance caused by the convergency difficulty; 2) models with the exponentially increased $\beta$ perform better than using linear increment.

\vspace{0.3em}
\noindent\textbf{Impact of $\theta$ and layer/head numbers.}\quad 
Hyperparameter $\theta$ tradeoffs the contribution of local temporal gradient information. As illustrated in Fig.~\ref{fig:dynamicloss}(b), PhysFormer could achieve smaller RMSE when $\theta$=0.4 and 0.7, indicating the importance of the normalized local temporal difference features for global spatio-temporal attention. We also investigate how the layer and head numbers influence the performance. As shown in Fig.~\ref{fig:layerhead}(a), with deeper temporal transformer blocks, the RMSE are reduced progressively despite heavier computational cost. In terms of the impact of head numbers, it is clear to find from Fig.~\ref{fig:layerhead}(b) that PhysFormer with four heads perform the best while fewer heads lead to sharp performance drops.    

\begin{figure}
\centering
\includegraphics[scale=0.13]{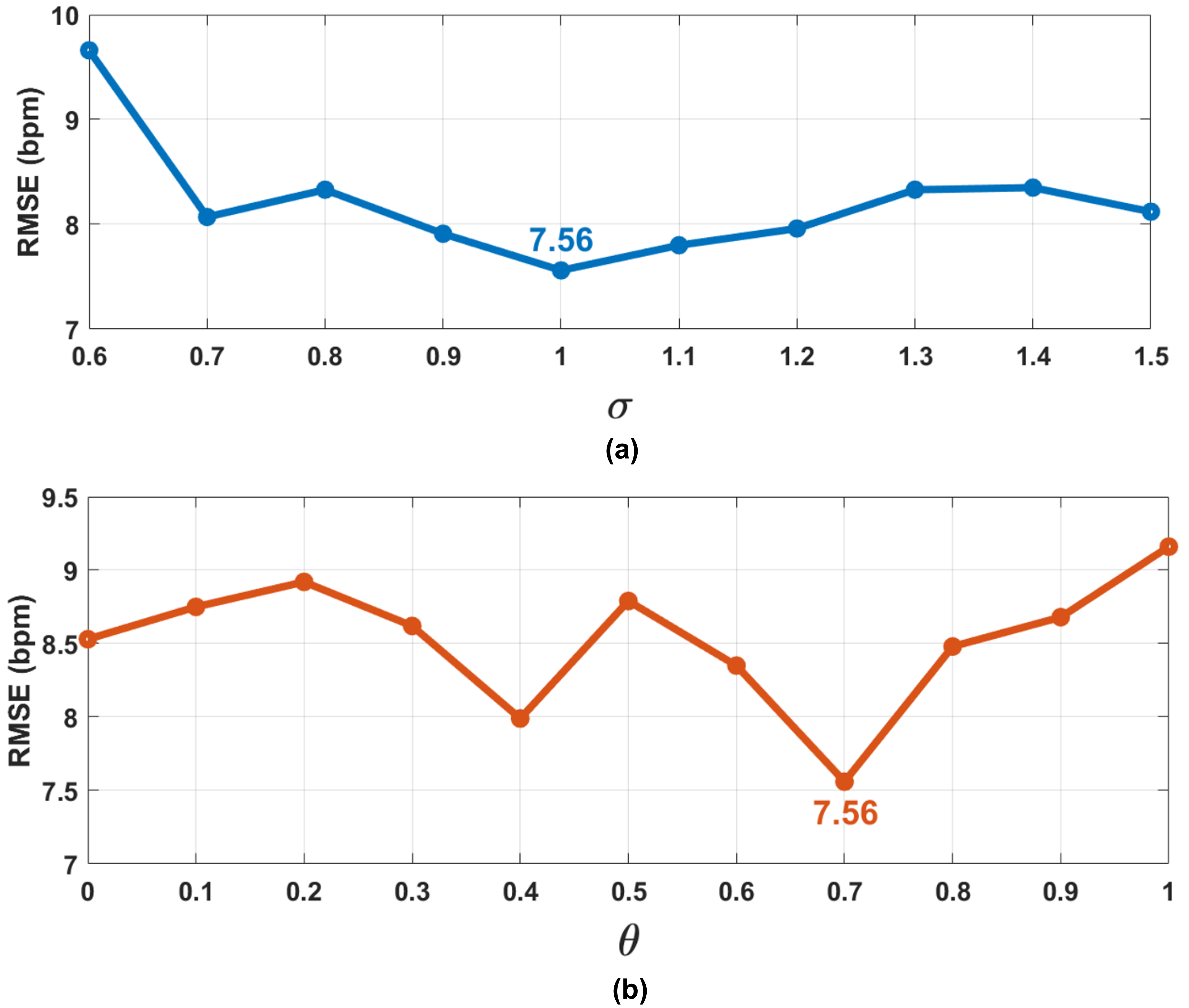}
\vspace{-0.8em}
  \caption{\small{
 Impacts of the (a) $\sigma$ in label distribution learning and (b) $\theta$ in TD-MHSA. }
  }
\label{fig:distribution}
\vspace{-0.8em}
\end{figure}

\begin{figure}
\includegraphics[scale=0.4]{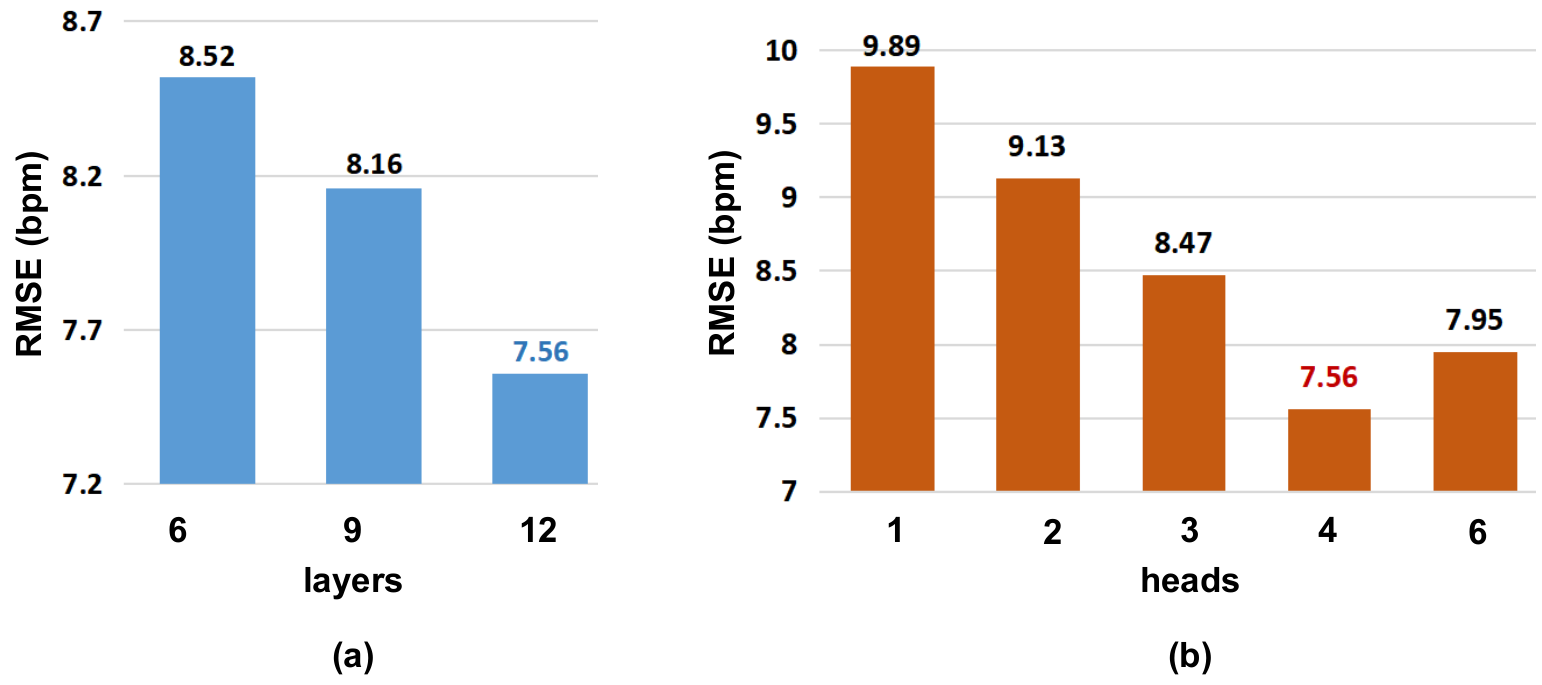}
\vspace{-1.1em}
\centering
  \caption{\small{
 Ablation of the (a) layers and (b) heads in PhysFormer. }
  }
\label{fig:layerhead}
\vspace{-1.6em}
\end{figure}

\subsection{Visualization and Discussion}
 \label{sec:Analysis}

\begin{figure}
\centering
\includegraphics[scale=0.24]{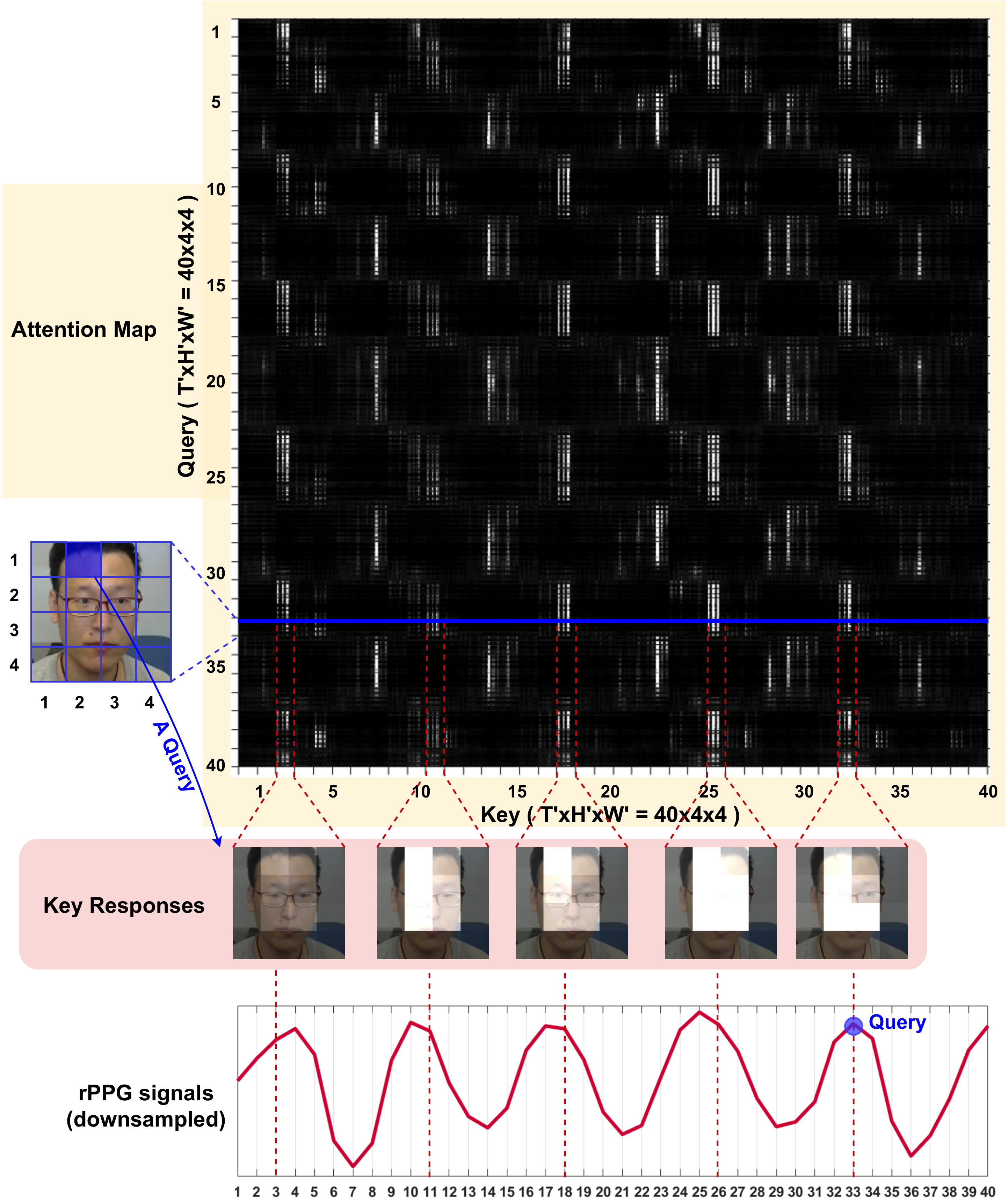}
\vspace{-0.9em}
  \caption{\small{
 Visualization of the attention map from the 1st head in last TD-MHSA module. Given the 530th tube token in \textcolor{blue}{blue} as a query, representative key responses are illustrated (the brighter, the more attentive). The predicted downsampled rPPG signals are shown for temporal attention understanding. }
  }
\label{fig:visualization}
\vspace{-0.8em}
\end{figure}

We visualize the attention map from the last TD-MHSA module as well as one example about the query-key interaction in Fig.~\ref{fig:visualization}. The x and y axes indicate the attention confidence from key and query tube tokens, respectively. From the attention map, we can easily find periodic or quasi-periodic responses along both axes, indicating the periodicity of the intrinsic rPPG features from PhysFormer. To be specific, given the 530th tube token (in blue) from the forehead (spatial face domain) and peak (temporal signal domain) locations as a query, the corresponding key responses are illustrated at the blue line in the attention map. On one hand, it can be seen from the key responses that dominant spatial attentions focus on the facial skin regions and discard unrelated background. On the other hand, the temporal localizations of the key responses are around peak positions in the predicted rPPG signals. All these patterns are reasonable: 1) the forehead and cheek regions~\cite{verkruysse2008remote} have richer blood volume for rPPG measurement and are also reliable since these regions are less affected by facial muscle movements due to e.g., facial expressions, talking; and 2) rPPG signals from healthy people are usually periodic. 

However, we also find two limitations of the spatio-temporal attention from Fig.~\ref{fig:visualization}. First, there are still some unexpected responses (e.g., continuous query tokens with similar key responses) in the attention map, which might introduce task-irrelevant noise and damage to the performance. Second, the temporal attentions are not always accurate, and some are coarse with phase shifts (e.g., the first vertical dotted line of the rPPG signals in bottom Fig.~\ref{fig:visualization}).

\vspace{-0.5em}
\section{Conclusions and Future Work}
\vspace{-0.1em}
\label{sec:conc}

In this paper, we propose an end-to-end video transformer architecture, namely PhysFormer, for remote physiological measurement. With temporal difference transformer and elaborate supervisions, PhysFormer is able to achieve superior performance on benchmark datasets. 
 The study of video transformer based physiological measurement is still at an early stage. Future directions include: 1) Designing more efficient architectures. The proposed PhysFormer is with 7.03 M parameters and 47.01 GFLOPs, which is unfriendly for mobile deployment; 2) Exploring more accurate yet efficient spatio-temporal self-attention mechanism especially for long sequence rPPG monitoring.


\noindent\textbf{Acknowledgment}  \quad This work was supported by the Academy of Finland for Academy Professor project EmotionAI (grants 336116, 345122), and ICT 2023 project (grant 328115),  by Ministry of Education and Culture of Finland for AI forum project, the National Natural Science Foundation of China (Grant No. 62002283), the EPSRC grant: Turing AI Fellowship: EP/W002981/1, EPSRC/MURI grant EP/N019474/1. We would also like to thank the Royal Academy of Engineering and FiveAI. As well, the authors wish to acknowledge CSC-IT Center for Science, Finland, for computational resources.

{\small
\bibliographystyle{ieee_fullname}
\bibliography{egbib}

\begin{thebibliography}{10}\itemsep=-1pt

\bibitem{arnab2021vivit}
Anurag Arnab, Mostafa Dehghani, Georg Heigold, Chen Sun, Mario Lu{\v{c}}i{\'c},
  and Cordelia Schmid.
\newblock Vivit: A video vision transformer.
\newblock {\em ICCV}, 2021.

\bibitem{bengio2009curriculum}
Yoshua Bengio, J{\'e}r{\^o}me Louradour, Ronan Collobert, and Jason Weston.
\newblock Curriculum learning.
\newblock In {\em ICML}, 2009.

\bibitem{bertasius2021space}
Gedas Bertasius, Heng Wang, and Lorenzo Torresani.
\newblock Is space-time attention all you need for video understanding?
\newblock {\em arXiv:2102.05095}, 2021.

\bibitem{bulat2021space}
Adrian Bulat, Juan-Manuel Perez-Rua, Swathikiran Sudhakaran, Brais Martinez,
  and Georgios Tzimiropoulos.
\newblock Space-time mixing attention for video transformer.
\newblock {\em NeurIPS}, 2021.

\bibitem{cao2021video}
Jiezhang Cao, Yawei Li, Kai Zhang, and Luc Van~Gool.
\newblock Video super-resolution transformer.
\newblock {\em arXiv:2106.06847}, 2021.

\bibitem{carion2020end}
Nicolas Carion, Francisco Massa, Gabriel Synnaeve, Nicolas Usunier, Alexander
  Kirillov, and Sergey Zagoruyko.
\newblock End-to-end object detection with transformers.
\newblock In {\em ECCV}, 2020.

\bibitem{carreira2017quo}
Joao Carreira and Andrew Zisserman.
\newblock Quo vadis, action recognition? a new model and the kinetics dataset.
\newblock In {\em CVPR}, 2017.

\bibitem{chen2021crossvit}
Chun-Fu Chen, Quanfu Fan, and Rameswar Panda.
\newblock Crossvit: Cross-attention multi-scale vision transformer for image
  classification.
\newblock {\em ICCV}, 2021.

\bibitem{chen2021aniformer}
Haoyu Chen, Hao Tang, Nicu Sebe, and Guoying Zhao.
\newblock Aniformer: Data-driven 3d animation with transformer.
\newblock {\em BMVC}, 2021.

\bibitem{chen2021geometry}
Haoyu Chen, Hao Tang, Zitong Yu, Nicu Sebe, and Guoying Zhao.
\newblock Geometry-contrastive transformer for generalized 3d pose transfer.
\newblock In {\em AAAI}, 2022.

\bibitem{chen2018deepphys}
Weixuan Chen and Daniel McDuff.
\newblock Deepphys: Video-based physiological measurement using convolutional
  attention networks.
\newblock In {\em ECCV}, 2018.

\bibitem{chen2018video}
Xun Chen, Juan Cheng, Rencheng Song, Yu Liu, Rabab Ward, and Z~Jane Wang.
\newblock Video-based heart rate measurement: Recent advances and future
  prospects.
\newblock {\em IEEE Transactions on Instrumentation and Measurement}, 2018.

\bibitem{de2013robust}
Gerard De~Haan and Vincent Jeanne.
\newblock Robust pulse rate from chrominance-based rppg.
\newblock {\em IEEE Transactions on Biomedical Engineering}, 2013.

\bibitem{ding2021hr}
Mingyu Ding, Xiaochen Lian, Linjie Yang, Peng Wang, Xiaojie Jin, Zhiwu Lu, and
  Ping Luo.
\newblock Hr-nas: Searching efficient high-resolution neural architectures with
  lightweight transformers.
\newblock In {\em CVPR}, 2021.

\bibitem{dosovitskiy2020image}
Alexey Dosovitskiy, Lucas Beyer, Alexander Kolesnikov, Dirk Weissenborn,
  Xiaohua Zhai, Thomas Unterthiner, Mostafa Dehghani, Matthias Minderer, Georg
  Heigold, Sylvain Gelly, et~al.
\newblock An image is worth 16x16 words: Transformers for image recognition at
  scale.
\newblock {\em ICLR}, 2021.

\bibitem{fan2021multiscale}
Haoqi Fan, Bo Xiong, Karttikeya Mangalam, Yanghao Li, Zhicheng Yan, Jitendra
  Malik, and Christoph Feichtenhofer.
\newblock Multiscale vision transformers.
\newblock {\em ICCV}, 2021.

\bibitem{gao2017deep}
Bin-Bin Gao, Chao Xing, Chen-Wei Xie, Jianxin Wu, and Xin Geng.
\newblock Deep label distribution learning with label ambiguity.
\newblock {\em IEEE TIP}, 2017.

\bibitem{gao2018age}
Bin-Bin Gao, Hong-Yu Zhou, Jianxin Wu, and Xin Geng.
\newblock Age estimation using expectation of label distribution learning.
\newblock In {\em IJCAI}, 2018.

\bibitem{geng2013facial}
Xin Geng, Chao Yin, and Zhi-Hua Zhou.
\newblock Facial age estimation by learning from label distributions.
\newblock {\em IEEE TPAMI}, 2013.

\bibitem{gideon2021way}
John Gideon and Simon Stent.
\newblock The way to my heart is through contrastive learning: Remote
  photoplethysmography from unlabelled video.
\newblock In {\em ICCV}, 2021.

\bibitem{girdhar2019video}
Rohit Girdhar, Joao Carreira, Carl Doersch, and Andrew Zisserman.
\newblock Video action transformer network.
\newblock In {\em CVPR}, 2019.

\bibitem{han2020survey}
Kai Han, Yunhe Wang, Hanting Chen, Xinghao Chen, Jianyuan Guo, Zhenhua Liu,
  Yehui Tang, An Xiao, Chunjing Xu, Yixing Xu, et~al.
\newblock A survey on visual transformer.
\newblock {\em arXiv:2012.12556}, 2020.

\bibitem{han2021transformer}
Kai Han, An Xiao, Enhua Wu, Jianyuan Guo, Chunjing Xu, and Yunhe Wang.
\newblock Transformer in transformer.
\newblock {\em arXiv:2103.00112}, 2021.

\bibitem{he2021transreid}
Shuting He, Hao Luo, Pichao Wang, Fan Wang, Hao Li, and Wei Jiang.
\newblock Transreid: Transformer-based object re-identification.
\newblock {\em ICCV}, 2021.

\bibitem{hsu2017deep}
Gee-Sern Hsu, ArulMurugan Ambikapathi, and Ming-Shiang Chen.
\newblock Deep learning with time-frequency representation for pulse estimation
  from facial videos.
\newblock In {\em IJCB}, 2017.

\bibitem{khan2021transformers}
Salman Khan, Muzammal Naseer, Munawar Hayat, Syed~Waqas Zamir, Fahad~Shahbaz
  Khan, and Mubarak Shah.
\newblock Transformers in vision: A survey.
\newblock {\em arXiv:2101.01169}, 2021.

\bibitem{lam2015robust}
Antony Lam and Yoshinori Kuno.
\newblock Robust heart rate measurement from video using select random patches.
\newblock In {\em ICCV}, 2015.

\bibitem{lee2020meta}
Eugene Lee, Evan Chen, and Chen-Yi Lee.
\newblock Meta-rppg: Remote heart rate estimation using a transductive
  meta-learner.
\newblock In {\em ECCV}, 2020.

\bibitem{li2018obf}
Xiaobai Li, Iman Alikhani, Jingang Shi, Tapio Seppanen, Juhani Junttila, Kirsi
  Majamaa-Voltti, Mikko Tulppo, and Guoying Zhao.
\newblock The obf database: A large face video database for remote
  physiological signal measurement and atrial fibrillation detection.
\newblock In {\em FG}, 2018.

\bibitem{li2014remote}
Xiaobai Li, Jie Chen, Guoying Zhao, and Matti Pietikainen.
\newblock Remote heart rate measurement from face videos under realistic
  situations.
\newblock In {\em CVPR}, 2014.

\bibitem{lin2019tsm}
Ji Lin, Chuang Gan, and Song Han.
\newblock Tsm: Temporal shift module for efficient video understanding.
\newblock In {\em CVPR}, 2019.

\bibitem{lin2021survey}
Tianyang Lin, Yuxin Wang, Xiangyang Liu, and Xipeng Qiu.
\newblock A survey of transformers.
\newblock {\em arXiv:2106.04554}, 2021.

\bibitem{liu2021fuseformer}
Rui Liu, Hanming Deng, Yangyi Huang, Xiaoyu Shi, Lewei Lu, Wenxiu Sun, Xiaogang
  Wang, Jifeng Dai, and Hongsheng Li.
\newblock Fuseformer: Fusing fine-grained information in transformers for video
  inpainting.
\newblock In {\em ICCV}, 2021.

\bibitem{liu2020multi}
Xin Liu, Josh Fromm, Shwetak Patel, and Daniel McDuff.
\newblock Multi-task temporal shift attention networks for on-device
  contactless vitals measurement.
\newblock {\em NeurIPS}, 2020.

\bibitem{liu2021efficientphys}
Xin Liu, Brian~L Hill, Ziheng Jiang, Shwetak Patel, and Daniel McDuff.
\newblock Efficientphys: Enabling simple, fast and accurate camera-based vitals
  measurement.
\newblock {\em arXiv:2110.04447}, 2021.

\bibitem{liu2021camera}
Xin Liu, Shwetak Patel, and Daniel McDuff.
\newblock Camera-based physiological sensing: Challenges and future directions.
\newblock {\em arXiv:2110.13362}, 2021.

\bibitem{liu2021end}
Xiaolong Liu, Qimeng Wang, Yao Hu, Xu Tang, Song Bai, and Xiang Bai.
\newblock End-to-end temporal action detection with transformer.
\newblock {\em arXiv:2106.10271}, 2021.

\bibitem{liu2021swin}
Ze Liu, Yutong Lin, Yue Cao, Han Hu, Yixuan Wei, Zheng Zhang, Stephen Lin, and
  Baining Guo.
\newblock Swin transformer: Hierarchical vision transformer using shifted
  windows.
\newblock {\em ICCV}, 2021.

\bibitem{liu2021video}
Ze Liu, Jia Ning, Yue Cao, Yixuan Wei, Zheng Zhang, Stephen Lin, and Han Hu.
\newblock Video swin transformer.
\newblock {\em arXiv:2106.13230}, 2021.

\bibitem{lu2021hr}
Hao Lu and Hu Han.
\newblock Nas-hr: Neural architecture search for heart rate estimation from
  face videos.
\newblock {\em Virtual Reality \& Intelligent Hardware}, 2021.

\bibitem{lu2021dual}
Hao Lu, Hu Han, and S~Kevin Zhou.
\newblock Dual-gan: Joint bvp and noise modeling for remote physiological
  measurement.
\newblock In {\em CVPR}, 2021.

\bibitem{neimark2021video}
Daniel Neimark, Omri Bar, Maya Zohar, and Dotan Asselmann.
\newblock Video transformer network.
\newblock {\em arXiv:2102.00719}, 2021.

\bibitem{niu2017continuous}
Xuesong Niu, Hu Han, Shiguang Shan, and Xilin Chen.
\newblock Continuous heart rate measurement from face: A robust rppg approach
  with distribution learning.
\newblock In {\em IJCB}, 2017.

\bibitem{niu2018synrhythm}
Xuesong Niu, Hu Han, Shiguang Shan, and Xilin Chen.
\newblock Synrhythm: Learning a deep heart rate estimator from general to
  specific.
\newblock In {\em ICPR}, 2018.

\bibitem{niu2019rhythmnet}
Xuesong Niu, Shiguang Shan, Hu Han, and Xilin Chen.
\newblock Rhythmnet: End-to-end heart rate estimation from face via
  spatial-temporal representation.
\newblock {\em IEEE TIP}, 2019.

\bibitem{niu2020video}
Xuesong Niu, Zitong Yu, Hu Han, Xiaobai Li, Shiguang Shan, and Guoying Zhao.
\newblock Video-based remote physiological measurement via cross-verified
  feature disentangling.
\newblock In {\em ECCV}, pages 295--310. Springer, 2020.

\bibitem{niu2019robust}
Xuesong Niu, Xingyuan Zhao, Hu Han, Abhijit Das, Antitza Dantcheva, Shiguang
  Shan, and Xilin Chen.
\newblock Robust remote heart rate estimation from face utilizing
  spatial-temporal attention.
\newblock In {\em FG}, 2019.

\bibitem{nowara2021benefit}
Ewa~M Nowara, Daniel McDuff, and Ashok Veeraraghavan.
\newblock The benefit of distraction: Denoising camera-based physiological
  measurements using inverse attention.
\newblock In {\em ICCV}, 2021.

\bibitem{poh2010advancements}
Ming-Zher Poh, Daniel McDuff, and Rosalind Picard.
\newblock Advancements in noncontact, multiparameter physiological measurements
  using a webcam.
\newblock {\em IEEE transactions on biomedical engineering}, 2010.

\bibitem{poh2010non}
Ming-Zher Poh, Daniel~J McDuff, and Rosalind~W Picard.
\newblock Non-contact, automated cardiac pulse measurements using video imaging
  and blind source separation.
\newblock {\em Optics express}, 2010.

\bibitem{qiu2018evm}
Ying Qiu, Yang Liu, Juan Arteaga-Falconi, Haiwei Dong, and Abdulmotaleb
  El~Saddik.
\newblock Evm-cnn: Real-time contactless heart rate estimation from facial
  video.
\newblock {\em IEEE TMM}, 2018.

\bibitem{soleymani2011multimodal}
Mohammad Soleymani, Jeroen Lichtenauer, Thierry Pun, and Maja Pantic.
\newblock A multimodal database for affect recognition and implicit tagging.
\newblock {\em IEEE transactions on affective computing}, 2011.

\bibitem{vspetlik2018visual}
Radim {\v{S}}petl{\'\i}k, Vojtech Franc, and Jir{\'\i} Matas.
\newblock Visual heart rate estimation with convolutional neural network.
\newblock In {\em BMVC}, 2018.

\bibitem{touvron2021training}
Hugo Touvron, Matthieu Cord, Matthijs Douze, Francisco Massa, Alexandre
  Sablayrolles, and Herv{\'e} J{\'e}gou.
\newblock Training data-efficient image transformers \& distillation through
  attention.
\newblock In {\em ICML}, 2021.

\bibitem{tulyakov2016self}
Sergey Tulyakov, Xavier Alameda-Pineda, Elisa Ricci, Lijun Yin, Jeffrey~F Cohn,
  and Nicu Sebe.
\newblock Self-adaptive matrix completion for heart rate estimation from face
  videos under realistic conditions.
\newblock In {\em CVPR}, 2016.

\bibitem{vaswani2017attention}
Ashish Vaswani, Noam Shazeer, Niki Parmar, Jakob Uszkoreit, Llion Jones,
  Aidan~N Gomez, {\L}ukasz Kaiser, and Illia Polosukhin.
\newblock Attention is all you need.
\newblock In {\em NIPS}, 2017.

\bibitem{verkruysse2008remote}
Wim Verkruysse, Lars~O Svaasand, and J~Stuart Nelson.
\newblock Remote plethysmographic imaging using ambient light.
\newblock {\em Optics express}, 2008.

\bibitem{wang2021temporal}
Lining Wang, Haosen Yang, Wenhao Wu, Hongxun Yao, and Hujie Huang.
\newblock Temporal action proposal generation with transformers.
\newblock {\em arXiv:2105.12043}, 2021.

\bibitem{wang2017algorithmic}
Wenjin Wang, Albertus~C den Brinker, Sander Stuijk, and Gerard de Haan.
\newblock Algorithmic principles of remote ppg.
\newblock {\em IEEE Transactions on Biomedical Engineering}, 2017.

\bibitem{wang2021pyramid}
Wenhai Wang, Enze Xie, Xiang Li, Deng-Ping Fan, Kaitao Song, Ding Liang, Tong
  Lu, Ping Luo, and Ling Shao.
\newblock Pyramid vision transformer: A versatile backbone for dense prediction
  without convolutions.
\newblock {\em ICCV}, 2021.

\bibitem{xiao2021early}
Tete Xiao, Mannat Singh, Eric Mintun, Trevor Darrell, Piotr Doll{\'a}r, and
  Ross Girshick.
\newblock Early convolutions help transformers see better.
\newblock {\em NeurIPS}, 2021.

\bibitem{xu2021long}
Mingze Xu, Yuanjun Xiong, Hao Chen, Xinyu Li, Wei Xia, Zhuowen Tu, and Stefano
  Soatto.
\newblock Long short-term transformer for online action detection.
\newblock {\em arXiv:2107.03377}, 2021.

\bibitem{yu2020autohr}
Zitong Yu, Xiaobai Li, Xuesong Niu, Jingang Shi, and Guoying Zhao.
\newblock Autohr: A strong end-to-end baseline for remote heart rate
  measurement with neural searching.
\newblock {\em IEEE SPL}, 2020.

\bibitem{yu2021transrppg}
Zitong Yu, Xiaobai Li, Pichao Wang, and Guoying Zhao.
\newblock Transrppg: Remote photoplethysmography transformer for 3d mask face
  presentation attack detection.
\newblock {\em IEEE SPL}, 2021.

\bibitem{yu2019remote1}
Zitong Yu, Xiaobai Li, and Guoying Zhao.
\newblock Remote photoplethysmograph signal measurement from facial videos
  using spatio-temporal networks.
\newblock In {\em BMVC}, 2019.

\bibitem{yu2021facial}
Zitong Yu, Xiaobai Li, and Guoying Zhao.
\newblock Facial-video-based physiological signal measurement: Recent advances
  and affective applications.
\newblock {\em IEEE Signal Processing Magazine}, 2021.

\bibitem{yu2019remote2}
Zitong Yu, Wei Peng, Xiaobai Li, Xiaopeng Hong, and Guoying Zhao.
\newblock Remote heart rate measurement from highly compressed facial videos:
  an end-to-end deep learning solution with video enhancement.
\newblock In {\em ICCV}, 2019.

\bibitem{yu2021deep}
Zitong Yu, Yunxiao Qin, Xiaobai Li, Chenxu Zhao, Zhen Lei, and Guoying Zhao.
\newblock Deep learning for face anti-spoofing: a survey.
\newblock {\em arXiv:2106.14948}, 2021.

\bibitem{yu2021searching}
Zitong Yu, Benjia Zhou, Jun Wan, Pichao Wang, Haoyu Chen, Xin Liu, Stan~Z Li,
  and Guoying Zhao.
\newblock Searching multi-rate and multi-modal temporal enhanced networks for
  gesture recognition.
\newblock {\em IEEE TIP}, 2021.

\bibitem{yuan2021tokens}
Li Yuan, Yunpeng Chen, Tao Wang, Weihao Yu, Yujun Shi, Zihang Jiang, Francis~EH
  Tay, Jiashi Feng, and Shuicheng Yan.
\newblock Tokens-to-token vit: Training vision transformers from scratch on
  imagenet.
\newblock {\em arXiv:2101.11986}, 2021.

\bibitem{zeng2020learning}
Yanhong Zeng, Jianlong Fu, and Hongyang Chao.
\newblock Learning joint spatial-temporal transformations for video inpainting.
\newblock In {\em ECCV}, 2020.

\bibitem{zhang2016joint}
Kaipeng Zhang, Zhanpeng Zhang, Zhifeng Li, and Yu Qiao.
\newblock Joint face detection and alignment using multitask cascaded
  convolutional networks.
\newblock {\em IEEE SPL}, 2016.

\bibitem{zhao2021tuber}
Jiaojiao Zhao, Xinyu Li, Chunhui Liu, Shuai Bing, Hao Chen, Cees~GM Snoek, and
  Joseph Tighe.
\newblock Tuber: Tube-transformer for action detection.
\newblock {\em arXiv:2104.00969}, 2021.

\bibitem{zheng2021rethinking}
Sixiao Zheng, Jiachen Lu, Hengshuang Zhao, Xiatian Zhu, Zekun Luo, Yabiao Wang,
  Yanwei Fu, Jianfeng Feng, Tao Xiang, Philip~HS Torr, et~al.
\newblock Rethinking semantic segmentation from a sequence-to-sequence
  perspective with transformers.
\newblock In {\em CVPR}, 2021.

\end{thebibliography}
}

\end{document}